\def\year{2022}\relax
\documentclass[letterpaper]{article} % DO NOT CHANGE THIS
\usepackage{aaai22}  % DO NOT CHANGE THIS
\usepackage{times}  % DO NOT CHANGE THIS
\usepackage{helvet}  % DO NOT CHANGE THIS
\usepackage{courier}  % DO NOT CHANGE THIS
\usepackage[hyphens]{url}  % DO NOT CHANGE THIS
\usepackage{graphicx} % DO NOT CHANGE THIS
\usepackage{natbib}  % DO NOT CHANGE THIS AND DO NOT ADD ANY OPTIONS TO IT
\usepackage{caption} % DO NOT CHANGE THIS AND DO NOT ADD ANY OPTIONS TO IT
\DeclareCaptionStyle{ruled}{labelfont=normalfont,labelsep=colon,strut=off} % DO NOT CHANGE THIS
\usepackage{algorithm}
\usepackage{algorithmic}

\usepackage{newfloat}
\usepackage{listings}
\floatstyle{ruled}
\newfloat{listing}{tb}{lst}{}
\floatname{listing}{Listing}
\author{
Anonymous AAAI 2022 Submission\\
Paper ID: 985
}
\usepackage{booktabs}
\usepackage{multirow}
\usepackage{adjustbox}
\usepackage{bm}
\usepackage{amsfonts}
\usepackage{graphicx}
\usepackage{amsthm}
\usepackage{amsmath}
\usepackage{amsmath}%
\usepackage{MnSymbol}%
\usepackage{wasysym}%
\usepackage{color}
\newcommand{\indep}{\perp \!\!\! \perp}

\theoremstyle{definition}

\newtheorem{proposition}{Proposition}

\newcommand{\tabincell}[2]{\begin{tabular}{@{}#1@{}}#2\end{tabular}}
\title{Invariant Information Bottleneck for Domain Generalization}

\author {
    Bo Li\textsuperscript{\rm 1},
    Yifei Shen\textsuperscript{\rm 2},
    Yezhen Wang\textsuperscript{\rm 1},
    \\
    Wenzhen Zhu\textsuperscript{\rm 3},
    Colorado Reed\textsuperscript{\rm 4},
    Kurt Keutzer\textsuperscript{\rm 4},
    \\
    Dongsheng Li\textsuperscript{\rm 1} and 
    Han Zhao\textsuperscript{\rm 5}
}
\affiliations {
    \textsuperscript{\rm 1} Microsoft Research Asia, China \\
    \textsuperscript{\rm 2} Hong Kong University of Science and Technology, China \\
    \textsuperscript{\rm 3} Washington University in St. Louis, USA\\
    \textsuperscript{\rm 4} University of California, Berkeley, USA \\
    \textsuperscript{\rm 5} University of Illinois at Urbana-Champaign, USA\\
}

\begin{document}

\maketitle
\begin{abstract}
Invariant risk minimization (IRM) has recently emerged as a promising alternative for domain generalization. Nevertheless, the loss function is difficult to optimize for nonlinear classifiers and the original optimization objective could fail when pseudo-invariant features and geometric skews exist. Inspired by IRM, in this paper we propose a novel formulation for domain generalization, dubbed invariant information bottleneck (IIB). IIB aims at minimizing invariant risks for nonlinear classifiers and simultaneously mitigating the impact of pseudo-invariant features and geometric skews. Specifically, we first present a novel formulation for invariant causal prediction via mutual information. Then we adopt the variational formulation of the mutual information to develop a tractable loss function for nonlinear classifiers. To overcome the failure modes of IRM, we propose to minimize the mutual information between the inputs and the corresponding representations. IIB significantly outperforms IRM on synthetic datasets, where the pseudo-invariant features and geometric skews occur, showing the effectiveness of proposed formulation in overcoming failure modes of IRM. Furthermore, experiments on DomainBed show that IIB outperforms $13$ baselines by $0.9\%$ on average across $7$ real datasets.
\end{abstract}

\section{Introduction}
In most statistical machine learning algorithms, a fundamental assumption is that the training data and test data are \emph{independently and identically distributed} (i.i.d.). However, the data we have in many real-world applications are not i.i.d. Distributional shifts are ubiquitous. Under such circumstances, classic statistical learning paradigms with strong generalization guarantees, e.g., Empirical Risk Minimization (ERM)~\cite{DBLP:journals/tnn/Vapnik99}, often fail to generalize due to the violation of the i.i.d.~assumption. It has been widely observed that the performance of a model often deteriorates dramatically when it is faced with samples from a different domain, even under a mild distributional shift~\cite{DBLP:journals/corr/irm}. On the other hand, collecting training samples from all possible future scenarios is essentially infeasible. Hence, understanding and improving the generalization of models on \emph{out-of-distribution} data is crucial.

\emph{Domain generalization} (DG), which aims to learn a model from several different domains so that it generalizes to \emph{unseen} related domains, has recently received much attention. From the perspective of representation learning, there are several paradigms towards this goal, including invariant representation learning~\cite{DBLP:conf/icml/MuandetBS13,DBLP:conf/nips/ZhaoZWMCG18,tachet2020domain}, invariant causality prediction~\cite{DBLP:journals/corr/irm,krueger2020out}, meta-learning \cite{DBLP:conf/nips/BalajiSC18,DBLP:conf/eccv/DuXXQZS020}, and feature disentanglement~\cite{DBLP:conf/eccv/DuXXQZS020,DBLP:conf/icml/PengHSS19}. Of particular interest is the invariant learning methods. Some early works, e.g., DANN \cite{DBLP:series/acvpr/GaninUAGLLML17}, CDANN \cite{long2018conditional}, aim at finding representations that are invariant across domains. Nevertheless, learning invariant representations fails for domain adaptation or generalization when the marginal label distributions change between source and target domains \cite{zhao2019learning}. Recently, Invariant Causal Prediction (ICP), and its follow-up Invariant Risk Minimization (IRM), have attracted much interest. ICP assumes that the data are generated according to a structural causal model (SCM)~\cite{DBLP:journals/jmlr/Pearl10}. The causal mechanism for the data generating process is the same across domains, while the \emph{interventions} can vary among different domains. Under such data generative assumptions, IRM~\cite{DBLP:journals/corr/irm} attempts to learn an optimal classifier that is invariant across domains. ICP then argues that under the SCM assumption, such a classifier can generalize across domains.

Despite the intuitive motivations, IRM falls short in several aspects. First, the proposed loss function in \cite{DBLP:journals/corr/irm} is difficult to optimize when the classifier is nonlinear. Furthermore, it has been shown that IRM fails when the pseudo-invariant features \cite{DBLP:journals/corr/abs-2010-05761} or geometric skews exist~\cite{nagarajan2020understanding}. Under such circumstances, the classifier will utilize both the causal and spurious features, leading to a violation of invariant causal prediction. To address the first issue, we propose an information-theoretical formulation of invariant causal prediction and adopt a variational approximation to ease the optimization procedure. To tackle the second issue, we emphasize that the use of pseudo-invariant features or geometric skews will inevitably increase the mutual information between the inputs and the representations. Thus, to mitigate the impact of pseudo-invariant features and geometric skews, we propose to constrain this mutual information, which naturally leads to a formulation of information bottleneck. Our empirical results show that the proposed approach can effectively improve the accuracy when the pseudo-invariant features and geometric skews exist. 

\paragraph{Contributions:} We propose a novel information-theoretic formulation for domain generalization, termed as invariant information bottleneck (IIB). IIB aims at minimizing invariant risks while at the same time mitigating the impact of pseudo-invariant features and geometric skews. Specifically, our contributions can be summarized as follows:

%a recent study empirically suggested that these methods only marginally outperform ERM in various open benchmarks \cite{DBLP:journals/corr/abs-2007-01434}. Furthermore, various conditions have been shown where the invariant learning methods fail, e.g., the pseudo-invariant feature \cite{DBLP:journals/corr/abs-2010-05761} and geometric skews \cite{nagarajan2020understanding}. Under such circumstances, the classifier will utilize both causal feature and transformations of spurious features, which leads to the failure in identifying causality. In this paper, based on the structure causal model, we propose a novel information-theoretic formulation of finding causality. Specifically, to achieve both high prediction accuracy and invariant causal prediction, we maximize the mutual information while constraining the mutual information between the label and domain conditioned on the representation. Furthermore, in the view of information, the utilizing of pseudo-invariant feature or geometric skews will increase the mutual information between the input and the representation. Thus, these issues can be addressed by constraining this mutual information. In summary, our work provides the following \textbf{contributions}: 

\textbf{(1)} We propose a novel formulation for invariant causal prediction via mutual information. We further adopt variational approximation to develop tractable loss functions for nonlinear classifiers.

\textbf{(2)} To mitigate the impact of pseudo-invariant features and geometric skews, inspired by the information bottleneck principle, we propose to constrain the mutual information between the inputs and the representations. The effectiveness is verified by the synthetic experiments of failure modes \cite{DBLP:journals/corr/abs-2106-06607,nagarajan2020understanding}, where IIB significantly improves the performance of IRM.

\textbf{(3)} Empirically, we analyze IIB's performance with extensive experiments on both synthetic and large-scale benchmarks. We show that IIB is able to eliminate the spurious information better than other existing DG methods, and achieves consistent improvements on 7 datasets by 0.7\% on DomainBed~\cite{DBLP:journals/corr/abs-2007-01434}.

\begin{figure*}[tb]
\centering
\includegraphics[width=0.95\textwidth]{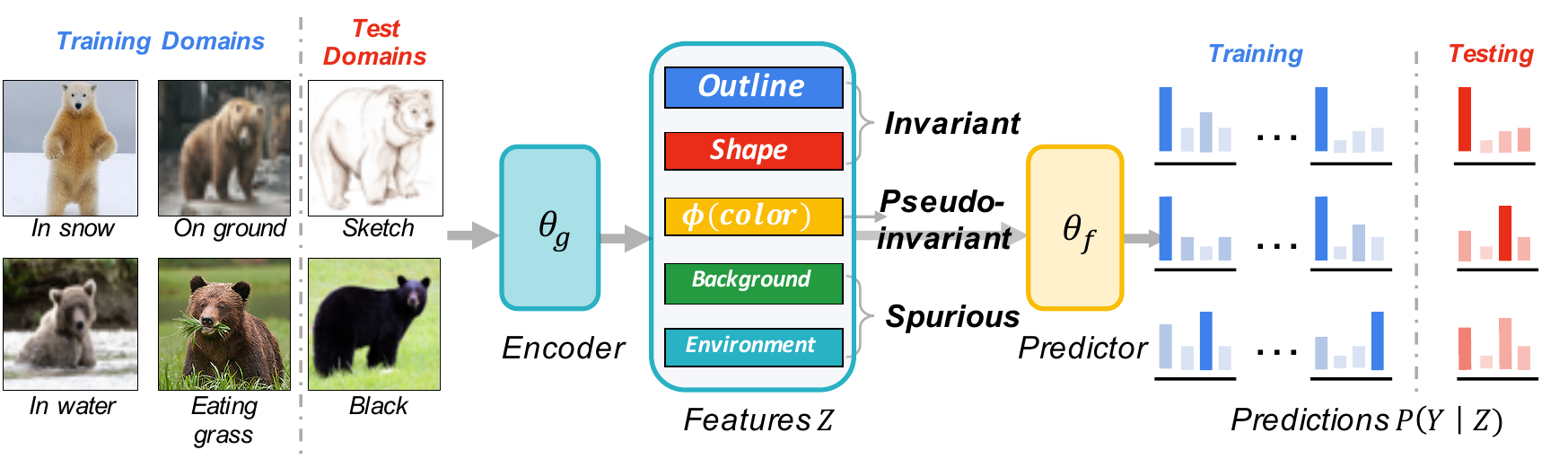}

\caption{Illustrations of features in OOD generalization. For all the bears in training domains, the predictions $P(Y \mid Z)$ conditioning on the invariant features (e.g. \emph{outline}) will be correct and invariant, while the predictions conditioning on pseudo-invariant features (possibly \emph{fur color} in this example) are misleading and may affect the generalization ability on test domains. Geometric skews~\citep{nagarajan2020understanding} are the spurious features used as a short-cut for max-margin classifiers. In this example, ERM will use all $5$ features as they are informative to labels. IRM, with the invariance constraint, will utilize the first $3$ features. IIB, by selecting the minimal sufficient features, only includes the shape or outline.}
% Images from training domains are all with different colors, the learned model would consider \emph{having a color} as invariant on existing data. However, this feature does not exist in test domains, hence conditioning on it would cause unpredicted errors during inference.
\label{fig:example}
\end{figure*}

\section{Related Work}
\subsection{Domain Generalization}
Existing methods of DG can be divided into three categories: (1)
\textbf{Data Manipulation}: Machine learning models typically rely on diverse training data to enhance the generalization ability. Data manipulation/augmentation methods~\cite{DBLP:conf/eccv/NazariK20,DBLP:conf/iclr/RiemerCALRTT19} aim to increase the diversity of existing training data with operations including flipping, rotation, etc. Domain randomization~\cite{DBLP:journals/corr/abs-1807-09834,DBLP:conf/iccv/YueZZSKG19,DBLP:conf/iccv/ZakharovKI19} provides more complex operations for image data, such as altering the location/texture of objects, replicating and resizing objects. In addition, there are some methods~\cite{DBLP:conf/iclr/RiemerCALRTT19,DBLP:conf/cvpr/QiaoZP20,DBLP:conf/nips/LiuLYW18,DBLP:journals/corr/abs-1905-12028,DBLP:conf/nips/ZhaoLYG0HCK19} that exploits generated data samples to enhance the model generalization ability. (2) \textbf{Ensemble Learning} methods~\cite{DBLP:conf/icip/ManciniBC018,DBLP:journals/corr/abs-2011-12672} assume that any sample in the test domain can be regarded as an integrated sample of the multiple-source domains, so the overall prediction should be inferred by a combination of the models trained on different domains. (3) \textbf{Meta-Learning} aims at learning a general model from multiple domains. In terms of domain generalization, MLDG~\cite{DBLP:conf/aaai/LiYSH18} divides data from the multiple domains into meta-train and meta-test to simulate the domain shift situation to learn the general representations. In particular, Meta-Reg~\cite{DBLP:conf/nips/BalajiSC18} learns a meta-regularizer for the classifier, and  Meta-VIB~\cite{DBLP:conf/eccv/DuXXQZS020} learns to generate the weights in the meta-learning paradigm by regularizing the KL divergence between marginal distributions of representations of the same category but from different domains. 

\subsection{Mutual Information-based Domain Adaptation}
Domain Adaptation is an important topic in the direction of transfer learning~\cite{long2015learning,ganin2016domain,tzeng2017adversarial,long2018conditional,zhao2021madan,zhao2020review,zhao2020multi,li2020rethinking}. The mutual information-based approaches have been widely applied in this area. The key idea is to learn a domain-invariant representation that are informative to the label, which can be formulated as \cite{DBLP:journals/corr/abs-2012-10713,DBLP:journals/corr/abs-2010-04647}
\begin{align} \label{eq:dann}
    \max_Z\quad I(Z,Y) - \lambda I(Z,A)
\end{align}
where $A$ is the identity of domains, $Z$ denotes the representation, and $Y$ denotes the labels. Commonly adopted implementations of \eqref{eq:dann} are DANN \cite{DBLP:series/acvpr/GaninUAGLLML17} and CDANN \cite{long2018conditional}. These implementations are also often adopted in domain generalization as baselines \cite{DBLP:journals/corr/abs-2007-01434}.

\subsection{Invariant Risk Minimization}
The above approaches enforces the invariance of the learned representations. On the other hand, Invariant Risk Minimization (IRM) suggests the invariance of feature-conditioned label distribution. Specifically, IRM seeks for an invariant causal prediction such that $\mathbb{E}[Y^e|\Phi(X^e)] = \mathbb{E}[Y^{e'}|\Phi(X^{e'})]$, for all $e,e' \in \mathcal{E}$. The objective of IRM is given by
\begin{align*}
    \min_{\bm{w}, \Phi } &\sum_{e \in \mathcal{E}_{\text{train} }} R^e(\bm{w} \circ \Phi), \\ \text{s.t. }  \bm{w}  \in & \ \underset{\hat{\bm{w}} }{\text{argmin}} \ R^e(\hat{\bm{w}} \circ \Phi),
\end{align*}
where $R^e$ is the cross-entropy loss for environment $e$, $\Phi$ is the feature extractor and $\bm{w}$ is a linear classifier. Note that the above objective is a bilevel optimization and difficult to optimize. Thus, in \cite{DBLP:journals/corr/irm}, first-order approximation is adopted and the loss function is given by 
\begin{equation} \label{eq:irm}
    \min_{\Phi} \ \sum_{e \in \mathcal{E}_{\text{train}} } R^e(\Phi) + \lambda \cdot \|\nabla_{w|w=1.0}  R^e(w \circ\Phi) \|,
\end{equation}
where $w \in \mathbb{R}$ is a dummy classifier.

\section{Preliminaries}
Failure modes of learning invariant representations are well-known in the literature~\citep{zhao2019learning,DBLP:journals/corr/abs-2012-10713}. Recently, some works have focused on characterizing the failure modes of IRM as well~\cite{DBLP:journals/corr/abs-2010-05761,nagarajan2020understanding}. As a motivation, we first briefly summarize these negative findings about IRM below.

\subsubsection{Pseudo-invariant Features} Even in the linear setting, it has been shown that the original IRM formulation~\eqref{eq:irm} cannot truly recover the features that induce invariant causal predictions~\cite{DBLP:journals/corr/abs-2010-05761}. Roughly speaking, in the linear case, one additional environment could be used to identify one spurious feature, and if the number of environments is smaller than the number of spurious features, some spurious features will leak to the algorithm-recovered causal features, which we call the \emph{pseudo-invariant features}. Specifically, we denote the causal features and spurious features as $\bm{z}_c$ and $\bm{z}_s$ respectively. According to the analysis in \cite{DBLP:journals/corr/abs-2010-05761}, there exists a transformation $\Phi$ such that $[\bm{z}_c, \Phi \bm{z}_s]$ are invariant features across the training dataset. Furthermore, the classifier will utilize $[\bm{z}_c, \Phi \bm{z}_s]$ instead of $\bm{z}_c$ to achieve a lower training error. The OOD generalization may fail due to the inclusion of $\bm{z}_s$, which can be arbitrary in the test dataset. An illustration of pseudo-invarinat features is shown in Fig.~\ref{fig:example}.

%The first problem is that the invariant features in the training dataset may not be the invariant features in the test dataset. 
\subsubsection{Geometric Skews}
\label{sec:gs} The OOD generalization can fail even if we assume the invariant features in the training dataset are also invariant in the test dataset due to the \emph{geometric skews}~\citep{nagarajan2020understanding}. It is observed that as the number of training points increase, the $\ell_2$-norm of the max-margin classifier grows. Specifically, we consider the case where an invariant feature $\bm{z}_{\text{inv} }$ is concatenated with a spurious feature $z_{\text{sp}}$ such that $\mathbb{P}[z_{\text{sp}} \cdot y > 0] > 0.5$. The dataset consists of a majority group $S_{\text{maj} }$ where $z_{\text{sp}} \cdot y > 0$ (e.g., cows/camels with green/yellow backgrounds) and a minority group $S_{\text{min} }$ where $z_{\text{sp}} \cdot y < 0$ (e.g., cows/camels with yellow/green backgrounds). Let $\bm{w}_{\text{all}}$ denote the least-norm classifier using invariant features to classify all samples and $\bm{w}_{\text{min}}$ denote the least-norm classifier using invariant features to classify the samples in $S_{\text{min} }$, and we have $\|\bm{w}_{\text{min}}\| \ll \|\bm{w}_{\text{all}}\|$. Hence, the algorithm can use the spurious feature as a short-cut to classify $S_{\text{maj}}$ and $S_{\text{min}}$, and then adopt $\bm{w}_{\text{min}}$ to classify the remaining $S_{\text{min}}$. This classifier using spurious feature will have a smaller norm than the invariant classifier, which leads to the failure of OOD generalization.

\begin{figure}[h]
\begin{center}
    \includegraphics[width=0.50\columnwidth]{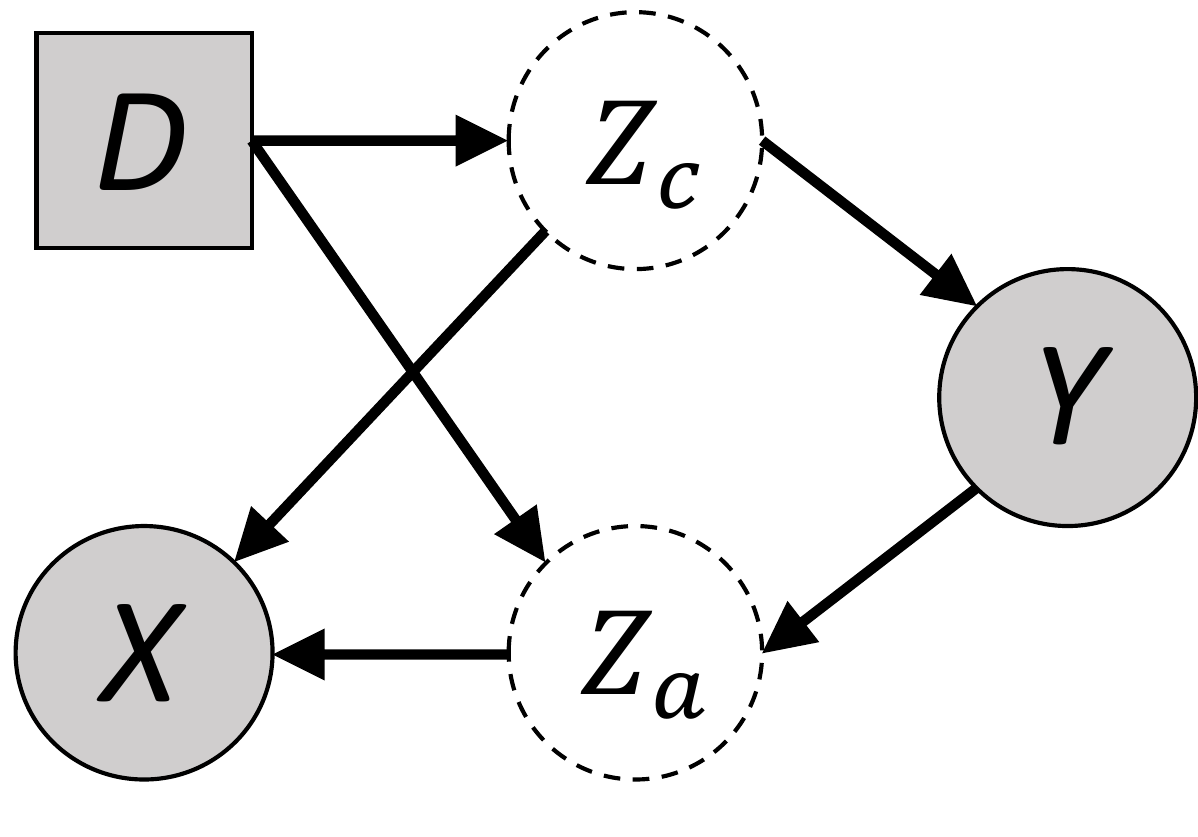}
\end{center}
\caption{A structural causal model explaining that different parts of an input $X$ have different causal relationships with the model output $Y$. Observed variables are shaded, while others are with dotted outlines.}
\label{fig:causal_graph}
\end{figure}

\section{Our Proposed Method}
In this section, we propose a novel information-theoretic objective of finding invariant causal relationship to overcome the two existing issues in the design of IRM objective. 

\subsection{Invariant Causal Prediction via Mutual Information}
Like other casual related works \cite{DBLP:conf/icml/ChangZYJ20,DBLP:journals/corr/abs-2006-07500}, we begin with a structural causal model, shown in Figure \ref{fig:causal_graph}. For simplicity, we leave out all the unnecessary elements. In general, we can see that an input $X$ can be divided into two variables, the \emph{causal} feature $Z_c$ and \emph{environmental} feature $Z_a$.  In Figure~\ref{fig:causal_graph}, we can readout that both features are correlated with $Y$, but only $Z_c$ is regarded as a causal feature. Through the concept of $d$-separation \cite{DBLP:journals/jmlr/Pearl10}, we can readout the conditional independence conditions that all data distributions $\mathcal{P}(D, X, Y)$ should satisfy:
\begin{enumerate}
    \item $Y \not\indep D$ means the marginal distribution of class label $Y$ can change across domains.
    \item $Y \indep D \mid Z_c$ means the class label $Y$ is independent of domain $D$ conditioned on the causal feature $Z_c$. The underlying causal mechanism determines that the value of $Y$ comes from its unique causal parent $Z_c$, which does not change across domains.
    \item $Y \not\indep D \mid Z_c, Z_a$ means that the conditional independence will not hold true if conditioned on both the \emph{causal} feature $Z_c$ and the \emph{environmental} features $Z_a$ since $Z_a$ is a collider between $D$ and $Y$. 
\end{enumerate}
The conditional independence tells us that only the real causal relation is stable and remains invariant across domains. In other words, we should eliminate the spurious environmental feature $Z_a$ by seeking the causal feature $Z_c$ that is independent of $D$ from $\Phi(X)$. Particularly, the representation $Z = \Phi(X)$ should have the following two merits: (1) $Z$ does not change among different domains for the same class label $Y$, hence achieving the conditional invariance of $Y \indep D \mid Z$; (2) $Z$ should be informative of the class label $Y$ (otherwise even a constant $\Phi(\cdot)$ would meet the first goal). The above two conditions coincide with the objective of IRM, and also suggest the following learning objective:
\begin{align}
    \label{eq:irm_information}
    \max_{\Phi} \ I(\Phi(X), Y) - \lambda I(Y, D \mid \Phi(X)),
\end{align}
where $\Phi$ is the feature extractor.

\begin{figure*}
    \centering
    \includegraphics[width=0.95\linewidth]{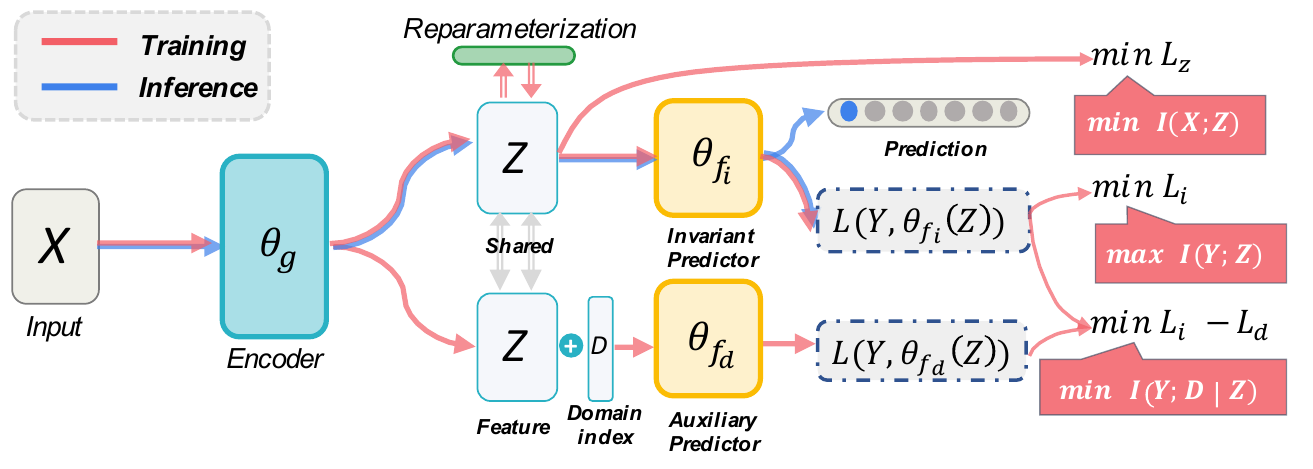}
    
    \caption{IIB optimizes a model consisting of three parts: (1) an invariant predictor $f_i(Z)$; (2) a domain-dependent predictor $f_d(Z, D)$; (3) an encoder $g(X)$. The three loss terms on the right hand side respectively correspond to the optimization of the three mutual information terms.}
    
    \label{fig:IIB_framework}
\end{figure*}

\begin{proposition}
Assume $I(Y,D|Z) = 0$, then we achieve invariant causal prediction in the sense that $\mathbb{E}[Y|\Phi(X) = x, D] = \mathbb{E}[Y|\Phi(X) = x]$.
\end{proposition}
\begin{proof}
Note that $I(Y,D|Z) = 0$ implies $Y$ and $D$ are independent conditioned on $\Phi(X)$. The conditional independence indicates that $\mathbb{P}(Y|\Phi(X) = x, D) = \mathbb{P}(Y|\Phi(X) = x)$, thus $\mathbb{E}[Y|\Phi(X) = x]$ is fixed and we can achieve invariant causal prediction.
\end{proof}

\subsection{On the Failure Modes of IRM}
\label{sec:ib}
In this subsection, we first scrutinize the failure conditions of IRM, i.e., pseudo-invariant features and geometric skews. Based on our analysis, among all the features that satisfy the invariant causal prediction constraint, we propose to use the one with the least capacity, i.e., the one that minimizes $I(X,Z)$. Alternatively, among all the feasible solutions, we are seeking the one that has the largest compression.

With pseudo-invariant features and geometric skews, the failure of existing approaches towards IRM is due to the inclusion of (transformations of) spurious features. We first give an example when the features are one-dimensional and the classifier is linear \cite{nagarajan2020understanding}. Denote the invariant feature, pseudo-invariant feature, feature causing geometric skews, spurious feature as $Z_i$, $Z_p$, $Z_{sk}$, and $Z_{sp}$. The overall features are $\bm{Z} = [Z_i, Z_p, Z_{sk}, Z_{sp}]$. In the ERM model, all the features will be adopted and OOD generalization fails. We consider the following optimization problem
\begin{align}
     \min_{w} & \sum_{e \in \mathcal{E}_{\text{train}  }} R^e(\bm{w} \cdot \bm{Z})\notag, \\
     \text{s.t. } & \|\bm{w}\|_0 \leq 1, \bm{w} \in  \underset{\hat{\bm{w}} }{\text{argmin}} \; R^e(\hat{\bm{w}} \cdot \bm{Z})
     ,\label{eq:sparse_invariant}
\end{align}
where $\|\bm{w}\|_0 \leq 1$ is the sparsity constraint, and $\bm{w} \in   {\text{argmin}}_{\hat{\bm{w}}} \; R^e(\hat{\bm{w}} \cdot \bm{Z})$ is the invariant risk constraint of IRM. Due to the sparsity constraint, there are only four choices. Choosing $Z_{sp}$ cannot satisfy the invariant constraint while choosing $Z_p$ or $Z_{sk}$ cannot minimize the empirical risk. Thus, the only optimal solution is $\bm{w} = [w_1^*, 0, 0, 0]$. Without the sparisty constraint, the optimization problem becomes IRM and $Z_i$, $Z_p$, $Z_{sk}$ will be used for classification. Without invariance constraint, $Z_{sp}$ might be chosen as the inclusion of spurious feature can lead to a lower empirical risk.

We then extend this intuition into the loss function design of deep neural networks in the view of mutual information. Suppose $Z_1, Z_2$ are features extracted from $X$, we have $I(X, [Z_1, Z_2]) \geq I(X,Z_1)$ as $Z_1$ is a subset of $[Z_1, Z_2]$. Thus, in order to select the one with the least capacity, we penalize a large $I(X,Z)$ by adding it to the original IRM formulation. To this end, we formulate our objective as 
\begin{align}
    \label{eq:iib_information}
    \max_{\Phi} I(\Phi(X), Y) - \lambda I(Y, D \mid \Phi(X)) - \beta I(X,\Phi(X)).
\end{align}
The term $I(Z, Y) - \beta I(X,Z)$ corresponds to the information bottleneck and $I(Y, D \mid Z)$ implements the IRM principle. As a result, we refer \eqref{eq:iib_information} as the \emph{invariant information bottleneck} (IIB) principle. 

\subsection{Loss Function Design}
The objective in \eqref{eq:iib_information} is still not a tractable loss function as the mutual information of high dimensional vectors is hard to estimate. Similar to VIB~\cite{DBLP:conf/iclr/AlemiFD017}, we leverage variational approximation to solve this issue. Let $r(z)$ be the  approximation to true marginal $p(z)$, and $q(y | z)$ to $p(y | z)$. Meanwhile let $p(z|x)$ be the stochastic encoder. Now the loss function of information bottleneck can be written as 
\begin{align} \label{eq:I(Z,Y)}
    I(Z, Y) &- \beta I(Z, X) \notag \\
    \geq \ \mathbb{E}_{p_{x, y, z }}\Big[ \log q(y \mid z)\Big]  &-  \beta \mathbb{E}_{p_{x, z}}\Big[ \log \frac{p(z \mid x)}{r(z)}\Big].
\end{align}
Optimizing \eqref{eq:I(Z,Y)} is still a difficult task. Then we transform it with reparametrization operation: We use an encoder of the form $p(z | x; g) = \mathcal{N}(z | g^{\mu}(x), g^{\Sigma}(x) )$, where $g$ outputs a $K$-dimensional mean $\mu$ of $z$ and a $K \times K$ covariance matrix $\Sigma$. Then by the change of variable formula we have $q(z | x)dz = q(\varepsilon)d\varepsilon$, where $z = g(x, \varepsilon)$, $\varepsilon \sim \mathcal{N}(0,1)$, so we can optimize \eqref{eq:I(Z,Y)} by optimizing
\begin{equation}
\label{eq:Lz_objective}
    \begin{aligned}
    \mathcal{L}_i (g, f_i) + \beta\mathcal{L}_z (g) ,
    % &   I(Y, D \mid Z) \\
    % = &\max_{q} \min_{r} \Exp_{x,y,d}[\log q(y | z, d) - \log r(y | z)] \\
    % = &\min_{r} \max_{q} \Exp_{x,y} [- \log r(y|g(x))] - \Exp_{x,y,d}[ - \log q(y | g(x), d) ] \\
    % % = & \min_{f_i} \max_{f_d} \Exp_{x,y,d} L(y, f_i(g(x))) - \Exp_{x,y,d} L(y, f_d(g(x), d)) \\
    % % = & \min_{f_i} \max_{f_d} \mathbb{E}_d\mathcal{R}^e(\Phi, f) - \mathbb{E}_d \mathcal{R}^e(\Phi, f_d)
    \end{aligned}
 \end{equation}
where $\mathcal{L}_i   = \min_{g, f_i}\mathbb{E}_{x, y}\Big[L(y, f_i(g(x)))\Big]$ and $\mathcal{L}_z  = \min_{g} \mathbb{E}_{x}\Big[KL[q(z|x; g) \| r(z)]\Big]$, where
%  \begin{align*}
%   \mathcal{L}_i  & = \min_{g, f_i}\mathbb{E}_{x, y}\big[L(y, f_i(g(x)))\big], \\
%      \mathcal{L}_z & = \min_{g} \mathbb{E}_{x}\big[\text{KL}[q(z|x; g) \| r(z)]\big],
%  \end{align*}
$g(x)$ is the feature extractor, $f_i$ is the classifier, and $L$ is the cross-entropy loss. 

We next proceed to deal with $I(Y,D \mid Z)$. Following the rules of variational approximation \cite{DBLP:conf/nips/FarniaT16}, we have 
% \int_{y \times z} p(y,z)
% \begin{equation}
\begin{align} \label{eq:I(Y,D|Z)}
    I(Y,D\mid Z) = & \ H(Y \mid Z) - H(Y \mid D, Z),
\end{align}
% \end{equation}
where $H(Y \mid Z) = - \sup_{q} \mathbb{E}_{p_{y, z}}\Big[\log q(y|z)\Big]$ and $H(Y \mid D, Z) = - \sup_{h} \mathbb{E}_{p_{y, z, d}} \Big[\log h(y \mid z,d)\Big]$.
Thanks to the universal approximation ability of neural networks, \eqref{eq:I(Y,D|Z)} can be written as the subtraction of two classification loss \cite{DBLP:conf/nips/FarniaT16}:

\begin{align} \label{eq:two_loss}
    I(Y,D\mid Z) \notag  = & \min_{f_i, g} \underbrace{\mathbb{E}_{x, y}\Big[L(y, f_i(g(x)))\Big]}_{\mathcal{L}_i} \\
    & - \min_{f_d, g} \underbrace{\mathbb{E}_{x, y, d}\Big[L(y, f_d(g(x), d))\Big]}_{\mathcal{L}_d},
\end{align}
where $f_i$ takes feature $z$ as the input, and $f_d, d = 1,\cdots,D$ takes both feature $z$ and domain index $d$ as the input. Overall, we can maximize our IIB objective function by optimizing its tractable lower bound:
  \begin{equation}
 \label{eq:full_objective}
    \begin{aligned}
    \min_{g, f_i} \max_{f_d} \ \mathcal{L}_i(g, f_i) + \beta \mathcal{L}_z (g) + \lambda \left(\mathcal{L}_i (g, f_i) - \mathcal{L}_d (g, f_d)\right).\notag
    % &   I(Y, D \mid Z) \\
    % = &\max_{q} \min_{r} \Exp_{x,y,d}[\log q(y | z, d) - \log r(y | z)] \\
    % = &\min_{r} \max_{q} \Exp_{x,y} [- \log r(y|g(x))] - \Exp_{x,y,d}[ - \log q(y | g(x), d) ] \\
    % % = & \min_{f_i} \max_{f_d} \Exp_{x,y,d} L(y, f_i(g(x))) - \Exp_{x,y,d} L(y, f_d(g(x), d)) \\
    % % = & \min_{f_i} \max_{f_d} \mathbb{E}_d\mathcal{R}^e(\Phi, f) - \mathbb{E}_d \mathcal{R}^e(\Phi, f_d)
    \end{aligned}
 \end{equation}
Guided by the above objective function, as illustrated in Figure~\ref{fig:IIB_framework}, IIB optimizes a model consisting of three parts: (1) an invariant predictor $f_i(Z)$; (2) an domain-dependent predictor $f_d(Z, D)$; (3) an encoder $g(X)$. The code implementation of IIB is released at Github.\footnote{\url{https://github.com/Luodian/IIB/tree/IIB}}
 
\begin{table}[tp]
\centering
\renewcommand{\arraystretch}{1.2}
\resizebox{\linewidth}{!}{%
\begin{tabular}{c|c|c}
\toprule
\textbf{Methods}  & \textbf{Validation Acc. (\%)} \bm{$\uparrow$} & \textbf{Test Acc.(\%)} \bm{$\uparrow$} \\ 
 \midrule[1.2pt]
% \midrule
ERM~\cite{DBLP:journals/tnn/Vapnik99} & 95.38 \scriptsize $\pm$ \scriptsize 0.03 & 11.16 \scriptsize $\pm$ \scriptsize 0.31 \\
% \hline
IRM~\cite{DBLP:journals/corr/irm} & 97.59 \scriptsize $\pm$ \scriptsize 1.39 & 57.98 \scriptsize $\pm$ \scriptsize 0.86 \\ 
% \hline
IB-ERM~\cite{DBLP:journals/corr/abs-2106-06607} & 97.64 \scriptsize $\pm$  \scriptsize 0.04 & 58.47 \scriptsize $\pm$ \scriptsize 0.86 \\ 
% \hline
IB-IRM~\cite{DBLP:journals/corr/abs-2106-06607} & 97.51 \scriptsize $\pm$ \scriptsize 1.09 & 71.79 \scriptsize $\pm$ \scriptsize 0.70 \\
\hline
% \midrule[1.1p]
IIB ($\lambda=0$) & 92.95 \scriptsize $\pm$ \scriptsize 0.50 & 69.52 \scriptsize $\pm$ \scriptsize 0.80 \\ 
% \hline
IIB ($\beta=0$) & 92.39 \scriptsize $\pm$ \scriptsize 0.50 & 66.93 \scriptsize $\pm$ \scriptsize 0.33 \\ 
% \hline
IIB & \textbf{98.11} \scriptsize $\pm$ \scriptsize 0.84 &  \textbf{74.23} \scriptsize $\pm$ \scriptsize 4.80 \\ 
\bottomrule
\end{tabular}%
}
\caption{Accuracy on CS-CMNIST experiment. We split 20\% from train set as validation set.}
\label{tab:ac_cs_mnist}
\end{table}

\section{Synthetic Experiments}
\subsection{Experimental Setup}
To validate IIB's efficacy of mitigating the impact of pseudo-invariant features and geometric skews, we adopt two types of synthetic experiments. Both pseudo-invariant features and geometric skews exist in the two experiments.

\paragraph{CS-CMNIST~\cite{DBLP:journals/corr/abs-2106-06607}}
CS-CMNIST is a ten-way classification task. The images are all drawn from MNIST. There are three environments, two training environments contain each 20,000 images, one test environment contains 20,000 images. There are ten colors associated with ten digit class correspondingly. The probability $p_e$ denotes that the image is colored with associated color. In two training environments, $p_e$ is set to 1 and 0.9, which means the images with certain class are colored with associated color with probability $p_e$ and are colored with random color with probability $1 - p_e$. In test environment, $p_e$ is set to 0, which means all images are colored at random. Overall, the color of images in training domains can be fully predictive to label with spurious features, i.e. using the associated color, but the information disappear at test domain. In CS-CMNIST, if the accuracy drops more at test time, it reflects that relying more on spurious features during training. We will give results of IIB on AC-CMNIST (in DomainBed it's known as CMNIST) in next section.

% The two experiments are to test the dependence of the algorithm on spurious features in OOD generalization setting; if the accuracy drops more at test time, the more it relies on spurious features during training. 

\begin{figure*}[tp]
\begin{center}
    \includegraphics[width=\textwidth]{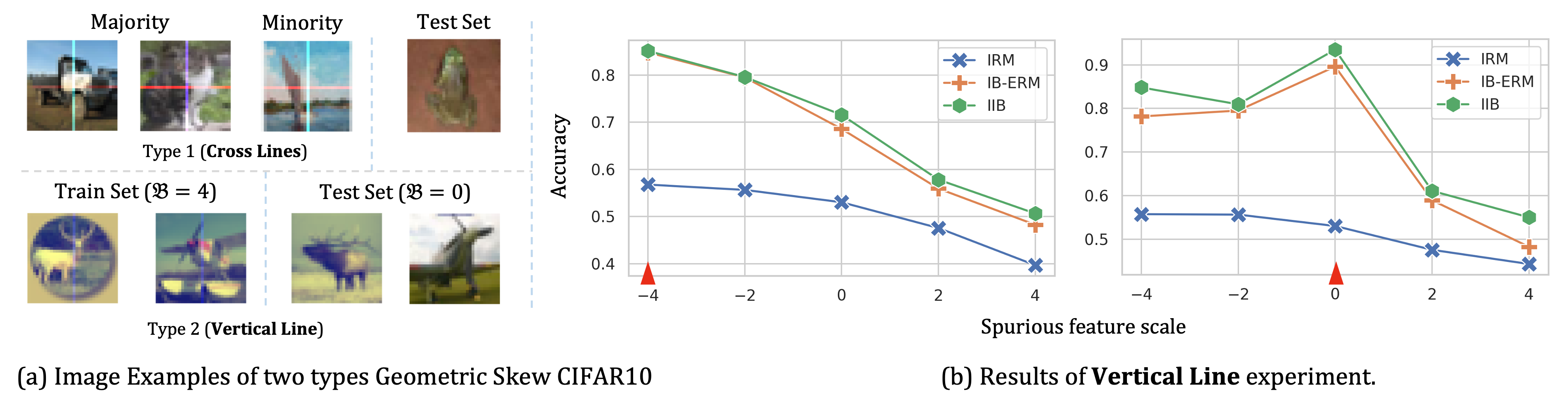}
\end{center}
\caption{(a) The above figure represents the image examples in majority/minority group in train set in Cross Lines experiment, while colored lines are not included in the test data. The below figure represents the image examples in train/test sets with different spurious feature scales in Vertical Line experiment. (b) Accuracy at test domains with different spurious feature scales $\mathcal{B}$. The upward-pointing red triangle denotes different $\mathcal{B}$ at training domains (we set them to -4 and 0 respectively). }
\label{fig:synthetic_dataset}
\end{figure*}

\begin{table}[tp]
\centering
\renewcommand{\arraystretch}{1.2}
\resizebox{\linewidth}{!}{%
\begin{tabular}{c|c|c}
\toprule
\textbf{Methods}  & \textbf{Validation Acc. (\%)} \bm{$\uparrow$} & \textbf{Test Acc.(\%)} \bm{$\uparrow$} \\
 \midrule[1.2pt]
% \hline
% \hline
ERM~\cite{DBLP:journals/tnn/Vapnik99} & 90.12 \scriptsize$\pm$ \scriptsize0.12 & 65.60 \scriptsize $\pm$ \scriptsize 0.27 \\
% \hline
IRM~\cite{DBLP:journals/corr/irm} & 63.82 \scriptsize$\pm$ \scriptsize0.25 & 42.68 \scriptsize $\pm$ \scriptsize 0.32 \\ 
% \hline
IB-ERM~\cite{DBLP:journals/corr/abs-2106-06607} & 83.93 \scriptsize$\pm$ \scriptsize0.10  & 69.70 \scriptsize $\pm$ \scriptsize 0.42 \\ 
% \hline
IB-IRM~\cite{DBLP:journals/corr/abs-2106-06607} & 81.61 \scriptsize$\pm$ \scriptsize0.69 & 65.82 \scriptsize$\pm$ \scriptsize0.77 \\
\hline
% \midrule[1.1p]
IIB ($\lambda=0$) & 79.97 \scriptsize $\pm$ \scriptsize 0.50 & 69.52 \scriptsize $\pm$ \scriptsize 0.80 \\ 
% \hline
IIB ($\beta=0$) & 78.47 \scriptsize $\pm$ \scriptsize 0.50 & 66.93 \scriptsize $\pm$ \scriptsize0.33 \\ 
% \hline
IIB & \textbf{92.86} \scriptsize$\pm$ \scriptsize0.29 & \textbf{71.04} \scriptsize $\pm$ \scriptsize 0.37 \\
\bottomrule
\end{tabular}%
}
\caption{Accuracy on Cross-Lines experiment. We split 20\% from train set as validation set.}
\label{tab:type_1_cross_lines}
\end{table}

\paragraph{Geometric Skew CIFAR10~\cite{nagarajan2020understanding}}
There are two types of tasks (as shown in Figure~\ref{fig:synthetic_dataset} (a)). For the first type, we name it \textit{Cross Lines} experiment, we create ten-valued spurious feature and add a vertical line passing through the middle of each channel, and also a horizontal line through the first channel. For these four lines added, we take the value of $(0.5 \pm 0.5 \mathcal{B})$ where $\mathcal{B} \in [-1, 1]$. Four lines, each with 2 choices, then we have a total of $2^4 = 16$ configurations. Among them, we choose the first 10 and denote the 10 configurations to each class in CIFAR10. For $i$-th configuration, corresponding to $i$-th class, we add this line with a probability of $p_{ii}=0.5$; for other $j$-th class, we set $p_{ij} = (1 - p_{ii})/10 = 0.05$. Taking the probability means $50\%$ data (the majority group) are correlated with spurious features (the specific colored line corresponding to each class), while other $5\%$ data (the minority group) are correlated with other 9 configurations at random. 
For the second type, we name it \textit{Vertical Line}, we add a colored line to the last channel of CIFAR10, regardless of the label during training, and vary its brightness during testing. In detail, we add a line with value choose from $\mathcal{B} \in [-4, 4]$. To avoid negative values, all pixels in last channel are added by 4, and then added by $\mathcal{B}$, and then divided by 9 to make sure the values lie in the range of $[0,1]$. Such an experiment would artificially create non-orthogonal components, where each data-point is represented on the plane of $(x_{\text{inv}}, x_{\text{inv}} + x_{\text{env}})$, rather than a more easy-to-disentangle representation under $(x_{\text{inv}}, x_{\text{env}})$. As discussed in~\cite{nagarajan2020understanding}, the model would be more susceptible to spurious features that may shift during testing.

\subsection{Observation for Results on Synthetic Experiments}
In CS-CMNIST, we compare IIB with several methods, including ERM~\cite{DBLP:journals/tnn/Vapnik99}, IRM, IB-IRM~\cite{DBLP:journals/corr/abs-2106-06607}. In particular, IB-IRM~\cite{DBLP:journals/corr/abs-2106-06607} is from a concurrent work, which proposes to combine information bottleneck and IRM to eliminate geometric skews. Among them (see Table~\ref{tab:ac_cs_mnist}), IIB has observable improvements over two synthetic datasets compared with other algorithms. Compare to IB-IRM, which is a direct combination of IB and IRM, our approach took a different approach to optimize the learning objective, which led to further enhancements. In the \textit{Cross Lines} experiment (see Table~\ref{tab:type_1_cross_lines}), we train the network on images with colored cross lines (each color corresponds to a specific class in CIFAR10), and test on normal images. From the improvements of IB over IRM, we observe that the information bottleneck structure can help mitigate the failure of IRM in geometric skews. In the \textit{Vertical Line} experiment (see Figure~\ref{fig:synthetic_dataset} (b)), we train the network on $\mathcal{B}$ = -4 or 0, and test on domains with different spurious feature scale $\mathcal{B}$. The results show that as the offset of spurious feature scale increases, the accuracy of training and testing environments decreases a lot. However, IIB still keeps good results even with large offset, indicating that it's effectiveness in alleviating the dependence on spurious feature. We have similar observations that information bottleneck (IB) could overcome the geometric skews which fails IRM.

% \usepackage{graphicx}

\begin{table*}[htp]
\centering %\scriptsize%\small
\setlength{\tabcolsep}{5pt}
\renewcommand{\arraystretch}{1.45}
\resizebox{\linewidth}{!}{%
\begin{tabular}{c |ccccccc| c}
\toprule
\small\textbf{Methods} &\textbf{Colored-MNIST}     & \textbf{Rotated-MNIST}     & \textbf{VLCS}             & \textbf{PACS}             & \textbf{\tabincell{c}{OfficeHome}}       & \textbf{\tabincell{c}{TerraIncognita}}   & \textbf{\tabincell{c}{DomainNet}}        & \textbf{Average}  \\
\midrule[1.2pt]
ERM~\cite{DBLP:journals/tnn/Vapnik99} & 36.7 \scriptsize $\pm$ \scriptsize 0.1& 97.7 \scriptsize $\pm$ \scriptsize 0.0            & 77.2 \scriptsize $\pm$ \scriptsize 0.4            & 83.0 \scriptsize $\pm$ \scriptsize 0.7            & 65.7 \scriptsize $\pm$ \scriptsize 0.5            & 41.4 \scriptsize $\pm$ \scriptsize 1.4            & 40.6 \scriptsize $\pm$ \scriptsize 0.2            & 63.2                      \\
% \hline
% CORAL~\cite{DBLP:conf/eccv/SunS16}                     & 39.7 \scriptsize $\pm$ \scriptsize 2.8            & 97.8 \scriptsize $\pm$ \scriptsize 0.1            & 78.7 \scriptsize $\pm$ \scriptsize 0.4            & 82.6 \scriptsize $\pm$ \scriptsize 0.5            & 68.5 \scriptsize $\pm$ \scriptsize 0.2            & 46.3 \scriptsize $\pm$ \scriptsize 1.7            & 41.1 \scriptsize $\pm$ \scriptsize 0.1            & 65.0                      \\
DANN~\cite{DBLP:series/acvpr/GaninUAGLLML17}                     & \textbf{40.7} \scriptsize $\pm$ \scriptsize 2.3            & 97.6 \scriptsize $\pm$ \scriptsize 0.2            & 76.9 \scriptsize $\pm$ \scriptsize 0.4            & 81.0 \scriptsize $\pm$ \scriptsize 1.1            & 64.9 \scriptsize $\pm$ \scriptsize 1.2            & 44.4 \scriptsize $\pm$ \scriptsize 1.1            & 38.2 \scriptsize $\pm$ \scriptsize 0.2            & 63.4                      \\
CDANN~\cite{DBLP:conf/eccv/LiTGLLZT18}                     & 39.1 \scriptsize $\pm$ \scriptsize 4.4            & 97.5 \scriptsize $\pm$ \scriptsize 0.2            & 77.5 \scriptsize $\pm$ \scriptsize 0.2            & 78.8 \scriptsize $\pm$ \scriptsize 2.2            & 64.3 \scriptsize $\pm$ \scriptsize 1.7            & 39.9 \scriptsize $\pm$ \scriptsize 3.2            & 38.0 \scriptsize $\pm$ \scriptsize 0.1            & 62.2                      \\
MLDG~\cite{DBLP:conf/aaai/LiYSH18}                     & 36.7 \scriptsize $\pm$ \scriptsize 0.2            & 97.6 \scriptsize $\pm$ \scriptsize 0.0            & 77.2 \scriptsize $\pm$ \scriptsize 0.9            & 82.9 \scriptsize $\pm$ \scriptsize 1.7            & 66.1 \scriptsize $\pm$ \scriptsize 0.5            & 46.2 \scriptsize $\pm$ \scriptsize 0.9            & 41.0 \scriptsize $\pm$ \scriptsize 0.2            & 64.0                      \\
IRM~\cite{DBLP:journals/corr/irm}                       & 40.3 \scriptsize $\pm$ \scriptsize 4.2            & 97.0 \scriptsize $\pm$ \scriptsize 0.2            & 76.3 \scriptsize $\pm$ \scriptsize 0.6            & 81.5 \scriptsize $\pm$ \scriptsize 0.8            & 64.3 \scriptsize $\pm$ \scriptsize 1.5            & 41.2 \scriptsize $\pm$ \scriptsize 3.6            & 33.5 \scriptsize $\pm$ \scriptsize 3.0            & 62.0                      \\
GroupDRO~\cite{DBLP:journals/corr/abs-1911-08731}                  & 36.8 \scriptsize $\pm$ \scriptsize 0.1            & 97.6 \scriptsize $\pm$ \scriptsize 0.1            & \textbf{77.9} \scriptsize $\pm$ \scriptsize 0.5            & 83.5 \scriptsize $\pm$ \scriptsize 0.2            & 65.2 \scriptsize $\pm$ \scriptsize 0.2            & 44.9 \scriptsize $\pm$ \scriptsize 1.4            & 33.0 \scriptsize $\pm$ \scriptsize 0.3            & 62.7                      \\
MMD~\cite{DBLP:conf/pkdd/AkuzawaIM19}                       & 36.8 \scriptsize $\pm$ \scriptsize 0.1            & 97.8 \scriptsize $\pm$ \scriptsize 0.1            & 77.3 \scriptsize $\pm$ \scriptsize 0.5            & 83.2 \scriptsize $\pm$ \scriptsize 0.2            & 60.2 \scriptsize $\pm$ \scriptsize 5.2            & 46.5 \scriptsize $\pm$ \scriptsize 1.5            & 23.4 \scriptsize $\pm$ \scriptsize 9.5            & 60.7                      \\
VREx~\cite{DBLP:journals/corr/abs-2003-00688} & 36.9 \scriptsize $\pm$ \scriptsize 0.3            & 93.6 \scriptsize $\pm$ \scriptsize 3.4            & 76.7 \scriptsize $\pm$ \scriptsize 1.0            & 81.3 \scriptsize $\pm$ \scriptsize 0.9            & 64.9 \scriptsize $\pm$ \scriptsize 1.3            & 37.3 \scriptsize $\pm$ \scriptsize 3.0            & 33.4 \scriptsize $\pm$ \scriptsize 3.1            & 60.6 \\
ARM~\cite{DBLP:journals/corr/abs-2007-02931}                       & 36.8 \scriptsize $\pm$ \scriptsize 0.0            & \textbf{98.1} \scriptsize $\pm$ \scriptsize 0.1            & 76.6 \scriptsize $\pm$ \scriptsize 0.5            & 81.7 \scriptsize $\pm$ \scriptsize 0.2            & 64.4 \scriptsize $\pm$ \scriptsize 0.2            & 42.6 \scriptsize $\pm$ \scriptsize 2.7            & 35.2 \scriptsize $\pm$ \scriptsize 0.1            & 62.2                      \\
Mixup~\cite{DBLP:journals/corr/abs-2001-00677}                     & 33.4 \scriptsize $\pm$ \scriptsize 4.7            & 97.8 \scriptsize $\pm$ \scriptsize 0.0            & 77.7 \scriptsize $\pm$ \scriptsize 0.6            & 83.2 \scriptsize $\pm$ \scriptsize 0.4            & 67.0 \scriptsize $\pm$ \scriptsize 0.2            & \textbf{48.7} \scriptsize $\pm$ \scriptsize 0.4            & 38.5 \scriptsize $\pm$ \scriptsize 0.3            & 63.8                      \\
% \hline
RSC~\cite{DBLP:conf/eccv/HuangWXH20}                       & 36.5 \scriptsize $\pm$ \scriptsize 0.2            & 97.6 \scriptsize $\pm$ \scriptsize 0.1            & 77.5 \scriptsize $\pm$ \scriptsize 0.5            & 82.6 \scriptsize $\pm$ \scriptsize 0.7            & 65.8 \scriptsize $\pm$ \scriptsize 0.7            & 40.0 \scriptsize $\pm$ \scriptsize 0.8            & 38.9 \scriptsize $\pm$ \scriptsize 0.5            & 62.7                      \\
MTL~\cite{DBLP:journals/jmlr/BlanchardDDLS21}                       & 35.0 \scriptsize $\pm$ \scriptsize 1.7            & 97.8 \scriptsize $\pm$ \scriptsize 0.1            & 76.6 \scriptsize $\pm$ \scriptsize 0.5            & 83.7 \scriptsize $\pm$ \scriptsize 0.4            & 65.7 \scriptsize $\pm$ \scriptsize 0.5            & 44.9 \scriptsize $\pm$ \scriptsize 1.2            & 40.6 \scriptsize $\pm$ \scriptsize 0.1            & 63.5                      \\
SagNet~\cite{nam2021reducing}                    & 36.5 \scriptsize $\pm$ \scriptsize 0.1            & 94.0 \scriptsize $\pm$ \scriptsize 3.0            & 77.5 \scriptsize $\pm$ \scriptsize 0.3            & 82.3 \scriptsize $\pm$ \scriptsize 0.1            & 67.6 \scriptsize $\pm$ \scriptsize 0.3            & 47.2 \scriptsize $\pm$ \scriptsize 0.9            & 40.2 \scriptsize $\pm$ \scriptsize 0.2            & 63.6                      \\
\hline
\textbf{IIB(Ours)} & 39.9 \scriptsize $\pm$ \scriptsize 1.2            & 97.2 \scriptsize $\pm$ \scriptsize 0.2            & 77.2 \scriptsize $\pm$ \scriptsize 1.6            & \textbf{83.9} \scriptsize $\pm$ \scriptsize 0.2            & \textbf{68.6} \scriptsize $\pm$ \scriptsize 0.1            & 45.8 \scriptsize $\pm$ \scriptsize  1.4 & \textbf{41.5} \scriptsize $\pm$ \scriptsize  2.3          &          \textbf{64.9}            \\
% \hline
% Deep Ensemble-4 networks & 99.36 & 0.99 & 0.08 & 92.40 & 1.80 & 0.26 & 73.58 & 1.30 & 0.82 & 51.28 & 2.40 & 1.81 \\ \hline
\bottomrule
\end{tabular}
}
\caption{%Performance comparison (Acc. \%) with different algorithms with \emph{leave one domain out} model selection strategy. Best accuracy is shown in boldface.
Performance comparison (Acc. \%) between the proposed IIB method and the state-of-the-art domain generalization methods with \emph{leave one domain out} model selection strategy. The best accuracy in each dataset is presented in boldface. The average accuracy over all the datasets is also reported.
}
\label{tab:4_main_benchmarks}
\end{table*}

\section{DomainBed Experiments}
To empirically corroborate the effectiveness of IIB, we conduct experiments on DomainBed~\cite{DBLP:journals/corr/abs-2007-01434} with 7 different datasets of different sizes.
% From small to large, we experiment on 7 different datasets of domain generalization task, including Colored-MINIST~\cite{DBLP:journals/corr/irm}, Rotated-MNIST~\cite{DBLP:conf/iccv/GhifaryKZB15}, PACS~\cite{DBLP:conf/iccv/LiYSH17}, VLCS~\cite{DBLP:conf/iccv/FangXR13}, Office-Home~\cite{DBLP:conf/cvpr/VenkateswaraECP17}, Terra Incognita~\cite{DBLP:conf/eccv/BeeryHP18}, DomainNet~\cite{DBLP:conf/iccv/PengBXHSW19}. 

% Furthermore, the domain partition is manually determined by the data collector. Clear domain partition gives the model stronger signals to distinguish data from different domains, whilst noisy domain partition creates harshness. There lies the difficulty of measuring an OOD Generalization model/algorithm. We hope that an algorithm can still perform well enough in the face of various data with different environmental noises (it may come from in domain or out of domain).

\paragraph{Model Selection Strategy} We choose two types of model selection strategies out of three in DomainBed. We do not test on the test-domain validation set, since it allows access to test domain while training. During training, the validation set is a subset of training set, we choose the model that performs best on the overall validation set for each domain. This strategy characterizes the in-distribution generalization capability of the model. In leave-one-domain-out cross validation, the training domains are separated from the test domain. This strategy characterizes the out-of-domain distribution generalization capacity of the model. Due to the space limit, we present results on leave-one-domain-out cross validation in Table~\ref{tab:4_main_benchmarks}, and put the results on training-domain validation set in supplementary materials.

\paragraph{Hyper-parameters and Implementation Details} In both selection strategies, for default hyper-parameters (e.g. learning rate, weight decay), we use default settings in DomainBed (e.g. learning rate is set to $1e-3$ for small images and with a selection range of $\textit{lr} \in [10^{-4.5}, 10^{-2.5}]$). For IIB specific hyper-parameters, we set $\lambda \in [1, 10^{2}]$, and $\beta \in [10^{-3}, 10^{-4}]$. For backbone feature extractor, in Rotated/Colored-MNIST, we use 4-layers 3x3 ConvNet. For VLCS and PACS, we use ResNet-18~\cite{DBLP:conf/eccv/HeZRS16}. For larger datasets, we opt to ResNet-50. For classifier, we both test linear and non-linear invariant (environment) classifiers. Specifically, in linear classifier, it has only one layer, otherwise it has three MLP layers with two RELU activation layers. For the increased number of parameters in the non-linear classifier, we correspondingly reduce the number of conv-layers in the backbone network to achieve a balance. We test the hyper-paramters and different model implementations on RotatedMNIST, the network is trained for 5000 iterations with batch size set to 128. We report the results in Table~\ref{tab:parameters}. We observe that the overall parameters under non-linear classifier setting are not increased too much. 

\subsection{Observation for Results on DomainBed}

From Table~\ref{tab:4_main_benchmarks}, we can see that IIB achieves the best \textit{average} performance on 7 datasets. On the other hand, the results in Table~\ref{tab:4_main_benchmarks} also show that there is no significant advantage of any domain generalization method that can dominate others in small datasets (Colored-MNSIT, Rotated-MNIST), which is consistent with the observations in~\citet{DBLP:journals/corr/abs-2007-01434}. IIB performs better than others in larger datasets (PACS, Office-Home, DomainNet), hence leading to a better average performance. We opine that the Information Bottleneck is able to better eliminate the noise from the spurious features in large datasets, while when the data set is small, this noise may still be useful as the short-cut in test domain for prediction, thus achieving better results.

\begin{table}[tp]
\centering
\renewcommand{\arraystretch}{1.2}
\resizebox{\linewidth}{!}{%
\begin{tabular}{c|c|c|c|c|c}
\toprule
\textbf{Classifier Type} & \textbf{MACs} & \textbf{Params} & \bm{$\beta$} & \bm{$\lambda$} & \textbf{Acc. (\%)} \bm{$\uparrow$}  \\
\midrule[1.2pt]
\multirow{4}{*}{linear} & \multirow{4}{*}{5.83G} & \multirow{4}{*}{370.95K} &1e-3 & 100 &  61.1 \\  \cline{4-6} 
 &  &  & \multirow{3}{*}{1e-4} & 1 &  94.7 \\ \cline{5-6} 
 &  &  & & 10 & 95.3 \\ \cline{5-6} 
 &  &  & & 100 &  95.1 \\ \hline
\multirow{4}{*}{non-linear} & \multirow{4}{*}{5.83G} & \multirow{4}{*}{375.33K} & 1e-3 & 100 &  63.2 \\ \cline{4-6} 
&  &  & \multirow{3}{*}{1e-4} & 1 &  96.8 \\ \cline{5-6} 
&  &  & & 10 &  97.2 \\ \cline{5-6} 
&  &  & & 100 &  \textbf{97.3} \\
\bottomrule
\end{tabular}
}
\caption{Different hyper-parameters' impact to the proposed IIB method on RotatedMNIST with leave-one-domain-out strategy. The results of multiply-add cumulation (MAC) operations and network parameters (Params) are reported. }
\label{tab:parameters}
\end{table}

\section{Conclusion}
Motivated by the existing limitations of the IRM methods for domain generalization, in this paper we developed a novel information-theoretic approach to overcome these issues. We term our new objective as the invariant information bottleneck (IIB). Our key insight in designing IIB lies in that when the number of training domains is not sufficient to identify all the potential spurious features, we should seek the ones that have the minimum capacity, among all the potential features that satisfy the original IRM objective. To implement IIB, we propose a variational approach to optimize the objective function that goes beyond the previous gradient penalty formulation, which only works for linear classifiers. The superior performance is demonstrated on both synthetic and real datasets through extensive experiments. As a future work, it is interesting to investigate the theoretical foundations of incorporating the information bottleneck principle in nonlinear invariant causal prediction and the effectiveness of IIB on regression tasks.
\clearpage
\begin{quote}
\begin{small}
\bibliography{aaai22}
\end{small}
\end{quote}
\end{document}

\maketitle
\section{Construction Details of Synthetic Dataset}
In this section, we will give more details on constructing the \textbf{Cross Line} and \textbf{Vertical Line} experiments. 
\subsection{Cross Lines Experiment}
Based on CIFAR10 dataset, we create ten-valued spurious feature and add a vertical line passing through the middle of each channel, and a horizontal line passing through the first channel. For each line added to the channel, we implement by adding the value taken of $0.5 \pm 0.5 \mathcal{B}$ where $\mathcal{B} \in [-1, 1]$. Four lines, each with two choices of $+\mathcal{B}$ or $-\mathcal{B}$. Then we have in total $2^4 = 16$ choices. We select 10 of the 16 configurations to map each configuration into one specific class. In detail, we select the images with specific class (e.g. bird), and add the the line with specific configuration (e.g. 0.5 + 0.5 $\mathcal{B}$, $\mathcal{B} = 0.8$). Similar with~\cite{nagarajan2020understanding}, we add the line of $i$-th configuration to corresponding class images with a probability of $p_{ii} = 0.5$; for other $j$-th class, we set $p_{ij} = (1-p_{ii}) / 10 = 0.05$. We call the $i$-th class images added with $i$-th configuration line with $p_{ii}$ the majority group. We call the $i$-th class images added with other $j$-th configuration line with $p_{ij}$ the minority group. The specific configurations are in the Table~\ref{tab:configuration}.

% \usepackage{multirow}
\setlength{\tabcolsep}{28pt}
\renewcommand{\arraystretch}{1.2}
\begin{table*}[tp]
\centering
\caption{10 configurations in \textbf{Cross Lines} experiments.}
\label{tab:configuration}
\begin{tabular}{|c|c|c|c|}
\hline
\textbf{Configuration \#} & \textbf{Channel} & \textbf{Line's Position} & \textbf{Sign}         \\ \hline
\multirow{4}{*}{0}        & 0                & Vertical                 & +  \\ \cline{2-4} 
                          & 1                & Vertical                 & +  \\ \cline{2-4} 
                          & 2                & Vertical                 & +  \\ \cline{2-4} 
                          & 0                & Horizontal               & +  \\ \hline
\multirow{4}{*}{1}        & 0                & Vertical                 & - \\ \cline{2-4} 
                          & 1                & Vertical                 & +  \\ \cline{2-4} 
                          & 2                & Vertical                 & +  \\ \cline{2-4} 
                          & 0                & Horizontal               & +  \\ \hline
\multirow{4}{*}{2}        & 0                & Vertical                 & +  \\ \cline{2-4} 
                          & 1                & Vertical                 & - \\ \cline{2-4} 
                          & 2                & Vertical                 & +  \\ \cline{2-4} 
                          & 0                & Horizontal               & +  \\ \hline
\multirow{4}{*}{3}        & 0                & Vertical                 & +  \\ \cline{2-4} 
                          & 1                & Vertical                 & +  \\ \cline{2-4} 
                          & 2                & Vertical                 & +  \\ \cline{2-4} 
                          & 0                & Horizontal               & +  \\ \hline
\multirow{4}{*}{4}        & 0                & Vertical                 & +  \\ \cline{2-4} 
                          & 1                & Vertical                 & +  \\ \cline{2-4} 
                          & 2                & Vertical                 & - \\ \cline{2-4} 
                          & 0                & Horizontal               & +  \\ \hline
\multirow{4}{*}{5}        & 0                & Vertical                 & +  \\ \cline{2-4} 
                          & 1                & Vertical                 & +  \\ \cline{2-4} 
                          & 2                & Vertical                 & +  \\ \cline{2-4} 
                          & 0                & Horizontal               & - \\ \hline
\multirow{4}{*}{6}        & 0                & Vertical                 & - \\ \cline{2-4} 
                          & 1                & Vertical                 & - \\ \cline{2-4} 
                          & 2                & Vertical                 & +  \\ \cline{2-4} 
                          & 0                & Horizontal               & +  \\ \hline
\multirow{4}{*}{7}        & 0                & Vertical                 & +  \\ \cline{2-4} 
                          & 1                & Vertical                 & - \\ \cline{2-4} 
                          & 2                & Vertical                 & - \\ \cline{2-4} 
                          & 0                & Horizontal               & +  \\ \hline
\multirow{4}{*}{8}        & 0                & Vertical                 & +  \\ \cline{2-4} 
                          & 1                & Vertical                 & +  \\ \cline{2-4} 
                          & 2                & Vertical                 & - \\ \cline{2-4} 
                          & 0                & Horizontal               & - \\ \hline
\multirow{4}{*}{9}        & 0                & Vertical                 & - \\ \cline{2-4} 
                          & 1                & Vertical                 & +  \\ \cline{2-4} 
                          & 2                & Vertical                 & +  \\ \cline{2-4} 
                          & 0                & Horizontal               & - \\ \hline
\multirow{4}{*}{10}       & 0                & Vertical                 & - \\ \cline{2-4} 
                          & 1                & Vertical                 & +  \\ \cline{2-4} 
                          & 2                & Vertical                 & - \\ \cline{2-4} 
                          & 0                & Horizontal               & +  \\ \hline
\end{tabular}
\end{table*}

\subsection{Vertical Line Experiment}
Based on CIFAR10 dataset, we add a vertical line to the last channel of all images. In detail, we add the line with value from $\mathcal{B} \in [-4, 4]$. To avoid negative values, all pixels in last channel are added by 4, and then added by $\mathcal{B}$, and then divided by 9 to ensure the pixels lie in the range of [0, 1]. Such operations will add non-orthogonal componenets to images, where each data-point is represented on the plane of ($x_{\text{inv}}, x_{\text{env}} + x_{\text{inv}}$) because of the added line with constant value in last channel. In~\cite{nagarajan2020understanding}, they show the non-orthogonal versus orthogonal experiment, the results suggest that the non-orthogonal images ($x_{\text{inv}}, x_{\text{env}} + x_{\text{inv}}$) are more hard-to-disentangle than orthogonal ones ($x_{\text{inv}}, x_{\text{env}}$), and would cause geometric skews of a max-margin classifier. In our experiment, during training, we set the $\mathcal{B}$ = 4 and 0, and test on domains with different $\mathcal{B} \in \{-4, -2, 0, 2, 4\}$. 

\section{More Results on DomainBed}
We report the rest experiment results in this section. In training-domain validation set, the validation set is subset of training set, we choose the model that performs best on the overall validation set for each domain. This strategy characterizes the in-distribution generalization capability of the model. 

The results are recorded in Table~\ref{tab:train_val}. From the Table, we can see that IIB achieves 67.5\% accuracy across 7 datasets on average, which is comparable to the best algorithm CORAL on DomainBed. Also IIB shows better performance on larger datasets (e.g. OfficeHome, DomainNet). The results demonstrate IIB's in-domains generalization ability.
\begin{table*}[htp]
\centering %\scriptsize%\small
\setlength{\tabcolsep}{5pt}
\renewcommand{\arraystretch}{1.45}
\caption{%Performance comparison (Acc. \%) with different algorithms with \emph{leave one domain out} model selection strategy. Best accuracy is shown in boldface.
Performance comparison (Acc. \%) between the proposed IIB method and the state-of-the-art domain generalization methods with \emph{training-domain validation set} model selection strategy. The best accuracy in each dataset is presented in boldface. The average accuracy over all the datasets is also reported.
}
\label{tab:train_val}
\resizebox{\linewidth}{!}{%
\begin{tabular}{c |ccccccc| c}
% \toprule
% \small\textbf{Methods} &\textbf{Colored-MNIST}     & \textbf{Rotated-MNIST}     & \textbf{VLCS}             & \textbf{PACS}             & \textbf{\tabincell{c}{OfficeHome}}       & \textbf{\tabincell{c}{TerraIncognita}}   & \textbf{\tabincell{c}{DomainNet}}        & \textbf{Average}  \\
% \midrule[1.2pt]

\toprule
\textbf{Algorithm}        & \textbf{ColoredMNIST}     & \textbf{RotatedMNIST}     & \textbf{VLCS}             & \textbf{PACS}             & \textbf{OfficeHome}       & \textbf{TerraIncognita}   & \textbf{DomainNet}        & \textbf{Avg}              \\
\midrule
ERM                       & 51.5 $\pm$ 0.1            & 98.0 $\pm$ 0.0            & 77.5 $\pm$ 0.4            & 85.5 $\pm$ 0.2            & 66.5 $\pm$ 0.3            & 46.1 $\pm$ 1.8            & 40.9 $\pm$ 0.1            & 66.6                      \\
IRM                       & 52.0 $\pm$ 0.1            & 97.7 $\pm$ 0.1            & 78.5 $\pm$ 0.5            & 83.5 $\pm$ 0.8            & 64.3 $\pm$ 2.2            & 47.6 $\pm$ 0.8            & 33.9 $\pm$ 2.8            & 65.4                      \\
GroupDRO                  & 52.1 $\pm$ 0.0            & 98.0 $\pm$ 0.0            & 76.7 $\pm$ 0.6            & 84.4 $\pm$ 0.8            & 66.0 $\pm$ 0.7            & 43.2 $\pm$ 1.1            & 33.3 $\pm$ 0.2            & 64.8                      \\
Mixup                     & 52.1 $\pm$ 0.2            & 98.0 $\pm$ 0.1            & 77.4 $\pm$ 0.6            & 84.6 $\pm$ 0.6            & 68.1 $\pm$ 0.3            & 47.9 $\pm$ 0.8            & 39.2 $\pm$ 0.1            & 66.7                      \\
MLDG                      & 51.5 $\pm$ 0.1            & 97.9 $\pm$ 0.0            & 77.2 $\pm$ 0.4            & 84.9 $\pm$ 1.0            & 66.8 $\pm$ 0.6            & 47.7 $\pm$ 0.9            & 41.2 $\pm$ 0.1            & 66.7                      \\
CORAL                     & 51.5 $\pm$ 0.1            & 98.0 $\pm$ 0.1            & \textbf{78.8} $\pm$ 0.6            & 86.2 $\pm$ 0.3            & 68.7 $\pm$ 0.3            & 47.6 $\pm$ 1.0            & 41.5 $\pm$ 0.1            & \textbf{67.5}   \\
MMD                       & 51.5 $\pm$ 0.2            & 97.9 $\pm$ 0.0            & 77.5 $\pm$ 0.9            & 84.6 $\pm$ 0.5            & 66.3 $\pm$ 0.1            & 42.2 $\pm$ 1.6            & 23.4 $\pm$ 9.5            & 63.3                      \\
DANN                      & 51.5 $\pm$ 0.3            & 97.8 $\pm$ 0.1            & 78.6 $\pm$ 0.4            & 83.6 $\pm$ 0.4            & 65.9 $\pm$ 0.6            & 46.7 $\pm$ 0.5            & 38.3 $\pm$ 0.1            & 66.1                      \\
CDANN                     & 51.7 $\pm$ 0.1            & 97.9 $\pm$ 0.1            & 77.5 $\pm$ 0.1            & 82.6 $\pm$ 0.9            & 65.8 $\pm$ 1.3            & 45.8 $\pm$ 1.6            & 38.3 $\pm$ 0.3            & 65.6                      \\
MTL                       & 51.4 $\pm$ 0.1            & 97.9 $\pm$ 0.0            & 77.2 $\pm$ 0.4            & 84.6 $\pm$ 0.5            & 66.4 $\pm$ 0.5            & 45.6 $\pm$ 1.2            & 40.6 $\pm$ 0.1            & 66.2                      \\
SagNet                    & 51.7 $\pm$ 0.0            & 98.0 $\pm$ 0.0            & 77.8 $\pm$ 0.5            & \textbf{86.3} $\pm$ 0.2            & 68.1 $\pm$ 0.1            & \textbf{48.6} $\pm$ 1.0            & 40.3 $\pm$ 0.1            & 67.2                      \\
ARM                       & \textbf{56.2} $\pm$ 0.2            & \textbf{98.2} $\pm$ 0.1            & 77.6 $\pm$ 0.3            & 85.1 $\pm$ 0.4            & 64.8 $\pm$ 0.3            & 45.5 $\pm$ 0.3            & 35.5 $\pm$ 0.2            & 66.1                      \\
VREx                      & 51.8 $\pm$ 0.1            & 97.9 $\pm$ 0.1            & 78.3 $\pm$ 0.2            & 84.9 $\pm$ 0.6            & 66.4 $\pm$ 0.6            & 46.4 $\pm$ 0.6            & 33.6 $\pm$ 2.9            & 65.6                      \\
RSC                       & 51.7 $\pm$ 0.2            & 97.6 $\pm$ 0.1            & 77.1 $\pm$ 0.5            & 85.2 $\pm$ 0.9            & 65.5 $\pm$ 0.9            & 46.6 $\pm$ 1.0            & 38.9 $\pm$ 0.5            & 66.1                      \\
\hline
\textbf{IIB(Ours)} & 52.0 $\pm$ 0.3 & 98.1 $\pm$ 0.2 & 77.6 $\pm$ 0.2 & 85.7 $\pm$ 0.6            & \textbf{69.0} $\pm$ 0.1 & 48.5 $\pm$ 0.4 & \textbf{41.6} $\pm$ 0.8 & \textbf{67.5} \\
\bottomrule
\end{tabular}
}
\end{table*}

\newpage

\begin{quote}
\begin{small}
\bibliography{aaai22}
\end{small}
\end{quote}